\documentclass[11pt]{article}

\usepackage[preprint]{acl}

\usepackage{times}
\usepackage{latexsym}
\usepackage[T1]{fontenc}
\usepackage[utf8]{inputenc}
\usepackage{microtype}
\usepackage{inconsolata}
\usepackage{graphicx}
\usepackage{booktabs}
\usepackage{amsmath}
\usepackage{multirow}
\usepackage{siunitx}
\usepackage{tikz}
\usetikzlibrary{positioning,arrows.meta,calc}
\usepackage{kotex}  % Korean (Hangul) for dataset examples; compile with XeLaTeX

\definecolor{oiblue}{HTML}{0072B2}
\definecolor{oigreen}{HTML}{009E73}
\definecolor{oigrey}{HTML}{999999}

\usepackage[capitalize]{cleveref}
\crefname{figure}{Figure}{Figures}\Crefname{figure}{Figure}{Figures}
\crefname{table}{Table}{Tables}\Crefname{table}{Table}{Tables}
\crefname{section}{Section}{Sections}\Crefname{section}{Section}{Sections}
\crefname{subsection}{Section}{Sections}\Crefname{subsection}{Section}{Sections}

\title{Reproducing and Stress-Testing Two Approaches to LLM Reasoning
Reliability: Test-Time Probability Aggregation and Logic-Representation Editing}

\author{
  Minhan Cho \\
  Remember \& Company AI Lab \\
  DSAIL, Dept.\ of Applied AI \\
  Sungkyunkwan University \\
  Seoul, Republic of Korea \\
  \texttt{zvezda@g.skku.edu}
  \And
  Jimin Kweon \\
  MAIN Lab, Dept.\ of Applied AI \\
  Sungkyunkwan University \\
  Seoul, Republic of Korea \\
  \texttt{kjm996@g.skku.edu}}

\begin{document}
\maketitle

%! PAPER: We independently reproduce two recent reliability methods and find that
%! RPC's test-time aggregation reproduces exactly but its new-domain edge over SC is
%! never significant, while LCF's logic-representation editing yields only a weak,
%! model-dependent effect that activation steering does not make model-agnostic.

%! PARA: We reproduce two reliability methods; RPC reproduces the grid but its
%! edge over SC is not significant on small sets, and LCF's one positive effect is
%! not significant either while it significantly reduces DeltaProb on two models.
\begin{abstract}
We independently reproduce two recent methods for making large language model (LLM) reasoning more reliable, and stress-test them across domains and models (RPC across four new task domains with Qwen3-8B, LCF across four $7$--$8$B models).
The first, \textbf{RPC}~\citep{zhou2025theoretical}, aggregates token probabilities and self-consistency at inference; the second, \textbf{LCF}~\citep{wu2025content}, trains projectors that split hidden states into ``content'' and ``logic'' and edits the logic part toward a valid region.
Validating such reliability claims matters because the original evaluations are run by each method's own authors and were never independently reproduced or stress-tested across models and domains, and LCF shipped no public code.
We re-run RPC's published-path aggregation and re-implement LCF's projector, contrastive, and intervention pipeline, then extend both to text-to-SQL, legal extraction, fallacy identification, and precedent grading, and probe LCF's representation directly.
RPC reproduces the original grid exactly on the authors' released reasoning paths; on four new domains its edge over self-consistency is never significant (ties or small mixed differences, paired $p\ge0.28$), and on BIRD, the one domain where we vary the budget, the edge grows with $K$ as predicted but its largest gap ($+2.5$ accuracy at $K{=}32$, $p{=}0.16$) reverses to $-0.25$ when we enlarge the sample to $n{=}200$. LCF's logic-validity direction is real but weak ($0.82$ separability at the single best sub-layer versus $0.95$ for a semantic-attribute control); its one positive effect (Qwen3 $\Delta$Prob) is not significant ($p{=}0.56$), while it significantly reduces $\Delta$Prob on two of the other three models.
\end{abstract}

\section{Introduction}
%! SECTION: Reliability methods are validated narrowly, so we reproduce two of
%! them and report where they hold and where they break.

\begin{figure*}[t]
\centering
\includegraphics[width=0.92\textwidth]{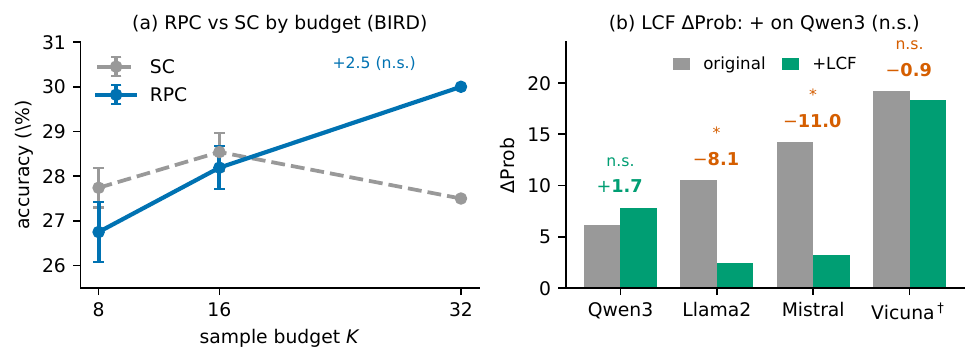}
\caption{\textbf{The two reliability methods diverge under faithful re-implementation.} (a) On BIRD, RPC's accuracy moves above SC (majority-vote baseline) only at $K{=}32$; error bars are 95\% CIs over $20$ seeds for $K{=}8,16$ ($K{=}32$ is a single deterministic run), and the $+2.5$ gap is not significant ($p{=}0.16$).
(b) $\Delta$Prob (the probability margin the model puts on the valid option) under LCF: it rises on Qwen3 (gap $+1.7$, n.s.) but \emph{significantly} degrades Llama-2 ($-8.1$) and Mistral ($-11.0$, both marked $*$); Vicuna ($-0.9$, n.s.) is an independent re-implementation re-run in fp16, so its scale is not comparable. The signed gap is printed above each pair.
$^\dagger$Vicuna uses a different validity judge, so its $\Delta$Prob is on a different scale (direction, not magnitude, is comparable).}
\label{fig:asymmetry}
\end{figure*}

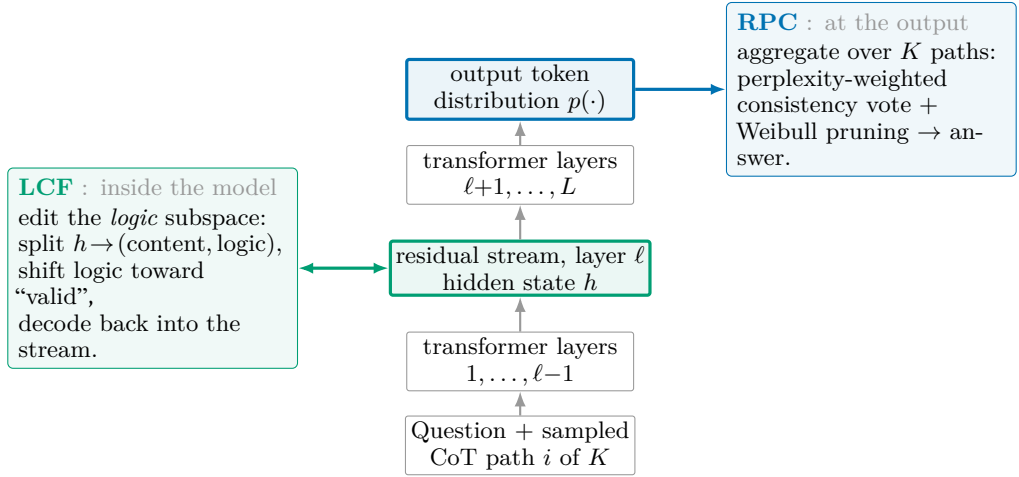
\begin{figure*}[t]
\centering
\begin{tikzpicture}[
  font=\small,
  stacklayer/.style={draw=oigrey, fill=white, rounded corners=1pt,
    minimum width=3.0cm, minimum height=0.62cm, align=center, inner sep=2pt},
  reslayer/.style={stacklayer, draw=oigreen, very thick, fill=oigreen!8},
  outlayer/.style={stacklayer, draw=oiblue, very thick, fill=oiblue!8},
  method/.style={draw, rounded corners=2pt, align=left, inner sep=4pt,
    text width=3.55cm, minimum height=1.55cm},
  uparrow/.style={-{Latex[length=2mm]}, oigrey, thick},
]
\node[stacklayer] (in) {Question $+$ sampled\\CoT path~$i$ of $K$};
\node[stacklayer, above=0.32cm of in] (low) {transformer layers\\$1,\dots,\ell{-}1$};
\node[reslayer,  above=0.46cm of low] (res) {residual stream, layer $\ell$\\\footnotesize hidden state $h$};
\node[stacklayer, above=0.46cm of res] (high) {transformer layers\\$\ell{+}1,\dots,L$};
\node[outlayer,  above=0.32cm of high] (out) {output token\\distribution $p(\cdot)$};
\foreach \a/\b in {in/low, low/res, res/high, high/out}
  \draw[uparrow] (\a) -- (\b);
\node[method, draw=oigreen, fill=oigreen!6, left=1.2cm of res, anchor=east] (lcf) {
  \textbf{\textcolor{oigreen}{LCF}}~\textcolor{oigrey}{\footnotesize: inside the model}\\[1pt]
  \footnotesize edit the \emph{logic} subspace:\\
  split $h\!\to\!(\text{content},\text{logic})$,\\
  shift logic toward ``valid'',\\
  decode back into the stream.};
\draw[{Latex[length=2mm]}-{Latex[length=2mm]}, oigreen, very thick] (lcf.east) -- (res.west);
\node[method, draw=oiblue, fill=oiblue!6, right=1.2cm of out, anchor=west] (rpc) {
  \textbf{\textcolor{oiblue}{RPC}}~\textcolor{oigrey}{\footnotesize: at the output}\\[1pt]
  \footnotesize aggregate over $K$ paths:\\
  perplexity-weighted\\
  consistency vote $+$\\
  Weibull pruning $\to$ answer.};
\draw[{Latex[length=2mm]}-, oiblue, very thick] (rpc.west) -- (out.east);
\end{tikzpicture}
\caption{\textbf{Where the two methods act.} Both consume the same sampled chain-of-thought but intervene at opposite ends of the model.
LCF (green) works \emph{inside} the network: at a single layer $\ell$ it splits the hidden state into a content and a logic vector, shifts the logic vector toward the valid region, and decodes it back into the residual stream.
RPC (blue) leaves the weights and activations untouched and instead aggregates the \emph{output} distributions of $K$ sampled paths with a perplexity-weighted consistency vote and a Weibull pruning step.
This contrast, representation editing versus output aggregation, organizes our reproduction.}
\label{fig:methods}
\end{figure*}

%! PARA: Reliability is defined, then two ways LLMs are unreliable that the
%! two papers target are named.
\emph{Reliability}, in the sense established by recent surveys, denotes a model that is accurate and also conveys trustworthy confidence in that accuracy: it is well calibrated, avoids confident factual errors, and judges an argument by its logical structure rather than its surface plausibility~\citep{geng2024survey, wang2023survey, dasgupta2022language}.
Large language models fall short on two of these axes that the methods we study target.
They calibrate sampled reasoning paths poorly, so selecting an answer from many samples is noisy~\citep{wang2022self, geng2024survey}; and they conflate whether a conclusion \emph{sounds} plausible with whether it \emph{follows} logically~\citep{wu2025content}.
Two recent papers attack these from opposite ends of the stack (\cref{fig:methods}).
RPC~\citep{zhou2025theoretical} stays at the output distribution and aggregates internal probability with self-consistency.
LCF~\citep{wu2025content} reaches into the residual stream and edits a learned ``logic'' subspace.
Both report strong reliability gains.\footnote{Code, data, and experiment logs: \url{https://github.com/rabqatab/llm-reasoning-reliability-reproduction}}

%! PARA: We ask whether the two methods reproduce and transfer, not whether they
%! are novel.
A reliability gain reported only by a method's own authors is a hypothesis, not a property.
We therefore ask two questions for each method: does it reproduce under a faithful re-implementation, and does the gain transfer across models and domains?
These questions are worth answering because neither method has been independently reproduced, and because LCF released no code, so any use of it requires a re-implementation whose fidelity is unknown.

%! PARA: We make four contributions: an RPC reproduction with four new domains and
%! a budget study, a from-scratch LCF analysis, a model-agnostic steering attempt,
%! and a per-method per-model account.
Our contributions are:
\begin{itemize}
  \item A faithful reproduction of RPC on the authors' published reasoning paths that matches their reported grid, plus four new evaluation domains and a sample-budget study isolating \emph{when} RPC beats self-consistency (\cref{sec:rpc}).
  \item A from-scratch re-implementation of LCF and a representation-level analysis showing the logic-validity signal is real but weak and not controllable by the additive intervention we tested, evaluated on four models including an independent cross-check (\cref{sec:lcf}).
  \item A literature-grounded attempt to make logic steering model-agnostic with contrastive activation steering~\citep{rimsky2024steering, valentino2026mitigating}, which fails on the fallacy task and succeeds only under narrow conditions on a controlled syllogism task (\cref{sec:agnostic}).
  \item A per-method, per-model account of strengths and weaknesses (\cref{sec:disc}).
\end{itemize}

%! PARA: The takeaway is an asymmetry: aggregation reproduces exactly and never
%! significantly hurts though its new-domain edge is n.s.,
%! while logic-representation editing does not transfer as a reliable intervention.
The takeaway is the asymmetry in \cref{fig:asymmetry}: test-time aggregation reproduces exactly and never significantly hurt any domain we tested, even though its edge over self-consistency on new domains does not reach significance, while representation editing for logic does not reproduce as a reliable, model-agnostic intervention.
On the present evidence a reader choosing between the two mechanisms has more reason to trust the former, though our small evaluation sets make this a directional rather than a decisive recommendation.

\section{Related Work}
%! SECTION: We position the two methods against the test-time scaling and
%! representation-editing literatures.

%! PARA: LLM reliability is framed by calibration and factuality literatures.
\paragraph{What reliability means for LLM reasoning.} A growing literature frames the reliability of an LLM as the alignment between its outputs and the trust a user can place in them, along several axes that trustworthiness benchmarks survey and measure together~\citep{huang2024position, wang2023decodingtrust}.
Calibration and confidence-estimation work asks whether a model's expressed or implicit confidence matches its accuracy, finding both that models can partly assess their own correctness~\citep{kadavath2022language} and that they remain systematically over-confident in many settings~\citep{geng2024survey}; prompting a model to state its confidence in words can be better calibrated than its raw token probabilities~\citep{tian2023just}.
Factuality and hallucination work asks whether generations are consistent with established facts~\citep{wang2023survey, huang2025survey}.
Reasoning reliability sits between the two: a fluent model can still mis-rank its own reasoning paths, or judge an argument by content rather than logical form~\citep{dasgupta2022language, lampinen2024language}, a gap that dedicated first-order-logic reasoning benchmarks expose~\citep{han2024folio}.
The two methods we reproduce target these failure modes directly, from the output distribution and from the residual stream respectively~\citep{zhou2025theoretical, wu2025content}.

%! PARA: RPC's lineage (CoT -> SC -> path-confidence), its evaluation gap, and adjacent critiques of vote- and confidence-based aggregation.
\paragraph{Test-time scaling, and RPC.} Chain-of-thought prompting made multi-step reasoning explicit, both few-shot~\citep{wei2022chain} and zero-shot~\citep{kojima2022large}.
Self-consistency~\citep{wang2022self} then sampled many chains and took a majority vote, trading compute for accuracy, and tree-structured search~\citep{yao2023tree}, compute-optimal allocation of test-time compute~\citep{snell2024scaling}, and process-level verifiers that score intermediate steps~\citep{lightman2024let} push this idea further.
Aggregating samples is not free of failure modes, though: majority voting can break on multi-step problems where step-level agreement does not track final correctness~\citep{chen2024two}.
A complementary line ranks a path by the model's own confidence in it, using generation likelihood or a stepwise self-evaluation signal rather than a majority vote~\citep{xie2023self}, while semantic-entropy methods cluster meaning-equivalent samples into a further confidence signal~\citep{farquhar2024detecting}.
RPC~\citep{zhou2025theoretical} unifies consistency and path-confidence signals~\citep{wang2022self, xie2023self} with a perplexity-weighted consistency vote plus a Weibull pruning step, and argues this speeds up convergence of the confidence estimate.
Its evaluation spans mathematics, code generation, and logical reasoning on a fixed set of open-source models, and reports aggregate task metrics (accuracy and expected calibration error)~\citep{zhou2025theoretical, guo2017calibration} without isolating \emph{when} the aggregation helps or whether it transfers to new models and to applied task domains such as text-to-SQL or legal extraction.
We reproduce its grid and then probe these gaps with four new task domains and a sample-budget sweep.
A concurrent analysis argues that probabilistic path confidence of this kind captures surface fluency more than logical structure~\citep{kim2026fluency}, offering one account of why the edge we measure stays small.

%! PARA: LCF's lineage (RepE -> ITI -> CAA) and its reproducibility gap.
\paragraph{Representation editing, and LCF.} A parallel line controls behavior by moving activations rather than re-sampling outputs: representation engineering~\citep{zou2023representation} reads and steers concept directions, inference-time intervention~\citep{li2023inference} shifts attention-head activations toward truthfulness, contrastive activation addition~\citep{rimsky2024steering, turner2023steering} adds a difference-of-means steering vector, sparse autoencoders extract interpretable feature directions from the residual stream~\citep{huben2024sparse}, and weight-space task arithmetic~\citep{ilharco2023task} composes fine-tuned directions.
LCF applies this idea to \emph{logical validity}, the very content-versus-logic distinction that LLMs are known to conflate~\citep{wu2025content, dasgupta2022language}: it trains projectors that disentangle a ``content'' and a ``logic'' subspace and edits the logic part toward a valid region.
A contemporaneous representational analysis likewise finds validity and plausibility to be linearly encoded and strongly aligned, and derives debiasing vectors that reduce content effects~\citep{bertolazzi2026validity}.
LCF, however, reports near-perfect control under an unreleased discriminator and ships no code~\citep{wu2025content}, and is validated on a fixed model set, so its central premise (a controllable logic direction) has not been independently checked.
We re-implement it from scratch and probe the representation directly.

%! PARA: Our approach: training-free conditional steering, and its open question.
\paragraph{Making the logic intervention model-agnostic.} Because a trained, model-specific projector is what makes LCF hard to reproduce, we ask whether a training-free steering recipe could replace it.
Conditional steering~\citep{lee2025programming} gates an activation edit on the input, and \citet{valentino2026mitigating} report that a kNN-conditional variant rescues otherwise-unresponsive models on a formal-reasoning task.
Mechanistic analyses of syllogistic inference find that models learn transferable, content-independent reasoning circuits that stay susceptible to belief bias~\citep{kim2024circuits}.
Those results are shown on their own curated syllogism data with a specific gate, so whether the recipe transfers to LCF's fallacy attribute is untested.
A concurrent result, moreover, shows that a decodable failure direction need not be correctable by a fixed linear edit when it overlaps task-critical computation~\citep{liu2026decodable}.
We take the contrastive-steering approach (a CAA direction with a conditional gate) and test it on the LCF fallacy task and on a matched-distribution syllogism task.
Methodologically the whole study is a reproducibility study: each published reliability gain is treated as a hypothesis to re-test under an independent implementation across models and domains, in the spirit of recent warnings about replication in language-model evaluation~\citep{vaugrante2024looming}.

\section{Datasets}
%! SECTION: Most of our evaluation data is constructed or re-processed by us, so
%! we document how each set was collected and built.
\label{sec:data}

\begin{table*}[t]
\centering
\small
\setlength{\tabcolsep}{3.5pt}
\begin{tabular}{l l p{3.0cm} p{4.1cm} l}
\toprule
Dataset & Used by & Source \& access & Construction in this work & Size \\
\midrule
Math (4 sets) & RPC & orig.\ benchmarks~\citep{hendrycks2021measuring, fang2025mathodyssey, he2024olympiadbench}; paths released by RPC authors (HF \texttt{WNJXYK}) & aggregated as released & paper grid \\
BIRD & RPC & \citet{li2023can}, \href{https://bird-bench.github.io}{bird-bench.github.io} & sample $K{\le}32$ SQL/q.\ (Qwen3), execution-match & 80 \\
KO-VER & RPC & Korean versioned legal citation extraction; in prep., not yet released & sample $K$ answers; score (law, article) set & 150 \\
KCC & RPC & \citet{cho2026kcc}, Korean civil precedents & balanced 4-class subset (80/grade), $K{=}16$ & 320 \\
LFUD & RPC, LCF & \citet{li2024reason}, \href{https://github.com/YandaGo/LFUD}{github.com/YandaGo/LFUD} & MCQ (RPC); valid/invalid conclusion pairs (LCF) & 540/204 \\
Legal-LCF & LCF & from KO-VER contexts & GPT-built valid/invalid conclusions & 140/40 \\
KCC-legal & LCF & from KCC holdings & GPT-built valid/invalid conclusions & 140/40 \\
Syllogism & LCF & synthetic (this work) & 30 triples $\to$ 4 cells, balanced 2$\times$2 & 80/40 \\
MoodRisk & probe & mental-health corpus & probe-only control (mean-pooled reps) & 8736 \\
\bottomrule
\end{tabular}
\caption{Datasets, and which method each supports.
The math paths, BIRD, and three added public benchmarks (GSM8K, FOLIO, LogiQA; \cref{tab:rpc-public}) are public; the rest are constructed or re-processed here.
``Size'' is evaluation items (train/test where both are used).
Per-dataset construction and one example each are in \cref{app:data}.}
\label{tab:datasets}
\end{table*}

%! PARA: Most of our data is built by us, and the three corpora a reader cannot
%! inspect are documented in the appendix with what we release in their place.
\cref{tab:datasets} lists every dataset.
Only the RPC math reasoning paths (sampled from four competition-mathematics benchmarks: MATH~\citep{hendrycks2021measuring}, MathOdyssey~\citep{fang2025mathodyssey}, OlympiadBench~\citep{he2024olympiadbench}, and the AIME competition) and the BIRD benchmark are taken as published.
The rest we collect, re-process, or synthesize, so we document each.
Three of them a reader cannot currently download: KO-VER and MoodRisk are under preparation by their originating projects, and KCC is released with its own paper~\citep{cho2026kcc}.
We use those under their projects' terms without redistributing their text, and \cref{app:kover} documents their provenance, label derivation, the exact slice we evaluate, and what we release in place of the corpora.

%! PARA: RPC extension datasets are sampled or re-processed by us.
\paragraph{RPC extensions.} For \textbf{BIRD}~\citep{li2023can} we sample up to $K{=}32$ SQL candidates per dev question with Qwen3-8B and score by execution match against the gold query, so the benchmark is public but the sampled paths are ours.
\textbf{KO-VER} pairs each sentence of a Korean court decision with the statutory citations that sentence makes; we sample $K$ extractions per item and score the normalized set of $(\text{law name}, \text{article})$ pairs, dropping the corpus's paragraph and statute-version fields (\cref{app:kover}).
\textbf{KCC}~\citep{cho2026kcc} is a Korean civil Supreme-Court precedent corpus of $2{,}939$ query--candidate pairs with \emph{graded} relevance labels $0$--$3$; we build a class-balanced subset of $80$ pairs per grade and generate Qwen3 grade predictions at $K{=}16$.
\textbf{LFUD}~\citep{li2024reason}, itself constructed with GPT-4, is a logical-fallacy corpus of $67$ propositions; for RPC we use its four-option fallacy-identification multiple-choice form.
Construction details and one worked example per dataset are in \cref{app:data}.

%! PARA: LCF datasets pair valid and invalid conclusions, generated for us.
\paragraph{LCF data.} LCF needs, per scenario, a premise with a logically \emph{valid} and a logically \emph{invalid} conclusion.
From LFUD's $67$ propositions (split $45{:}5{:}17$) we build the conclusion-generation set by taking the dataset's fallacious conclusion as the invalid member and generating a valid conclusion with GPT-4o-mini, and we reuse the four-option items for fallacy identification.
To test generalization beyond fallacies we build two independent \textbf{legal} analogues with the same schema: one whose premise is a KO-VER decision context (truncated to $1{,}200$ characters), and one whose premise is a KCC precedent holding; two separately sourced legal sets let us check that any effect is not a single-corpus artifact.

%! PARA: The syllogism and control sets are purpose-built.
\paragraph{Synthetic and control sets.} For the model-agnostic study (\cref{sec:agnostic}) we synthesize a \textbf{syllogism} set that crosses formal validity with believability: from $30$ ordered term-triples (subset relations such as roses~$\subset$~flowers~$\subset$~plants) we instantiate the four cells (valid/invalid $\times$ believable/unbelievable), balanced, split $80/40$.
The \textbf{MoodRisk} mental-health corpus ($8{,}736$ annotated posts with released layer embeddings) is used only as a control: a suicide-risk probe that bounds how strongly a \emph{semantic} attribute is linearly encoded, for comparison with the logic-validity probe.

\section{Experiments}
%! SECTION: We describe the two methods under test and how we re-implement and
%! evaluate them.
\label{sec:exp}

%! PARA: RPC combines perplexity-consistency with reasoning pruning.
\paragraph{Method under test: RPC.} Given $K$ sampled chain-of-thought paths~\citep{wei2022chain}, self-consistency (SC) picks the majority answer and perplexity (PPL) weights each path by its mean token probability.
RPC combines the two: \emph{Perplexity Consistency} weights consistency votes by path probability, and \emph{Reasoning Pruning} fits a Weibull mixture over path probabilities and discards a low-probability component.
RPC trains nothing; it is a drop-in aggregation rule over samples.

%! PARA: LCF trains projectors and edits the logic subspace.
\paragraph{Method under test: LCF.} LCF freezes the base LLM and trains two projectors that map a hidden state into a content vector and a logic vector, a cross-attention decoder that reconstructs the hidden state, and an InfoNCE contrastive objective that pulls logic vectors of valid conclusions together and pushes invalid ones away.
At inference it shifts the logic vector toward the valid region with a scale $\eta$ and decodes back into the residual stream.
The premise is that logic and content are separable directions and that moving along the logic direction steers logical behavior without changing content.

%! PARA: RPC reuses the official evaluator; LCF is built from scratch.
\paragraph{Re-implementation.} For RPC we reuse the authors' released aggregation and evaluation code over their published reasoning paths, and add domain-specific answer-equality functions for the new datasets; the exact per-dataset scoring functions, seeds, splits, and generation prompts are in the released code and \cref{app:data}.
LCF has no public code, so we re-implement the full pipeline: projectors ($d\!\to\!2048\!\to\!1024$), the cross-attention decoder, the InfoNCE loss ($\tau{=}0.1$), and the $\eta$-scaled intervention, following the paper's description.
Because the paper's self-trained validity discriminator is unreleased, we substitute an auditable GPT-4 judge and a DistilBERT validity classifier, and we flag every place this substitution affects a number.

%! PARA: Models, baselines, metrics, and hardware.
\paragraph{Models, baselines, and metrics.} We run Qwen3-8B~\citep{team2025qwen3} locally, and the paper's models Llama-2-7b-chat~\citep{touvron2023llama}, Mistral-7B-Instruct~\citep{jiang2023mistral}, and Vicuna-7b~\citep{chiang2023vicuna}.
RPC is compared against its two constituent baselines: \textbf{SC} (self-consistency, the majority vote over sampled answers) and \textbf{PPL} (perplexity, which selects the answer carried by the lowest-perplexity path).
For RPC we report accuracy and expected calibration error~\citep[ECE;][]{guo2017calibration}, averaged over $10$ seeds.
For LCF we report fallacy-identification accuracy and $\Delta$Prob (the probability margin the model places on the valid option), and conclusion-generation validity.
All experiments run on two DGX Spark GB10 nodes; the heaviest job is an 8B forward pass, so compute is modest.

\section{Results}
%! SECTION: RPC reproduces and is precondition-gated; LCF's logic editing is weak,
%! model-dependent, and not made model-agnostic by activation steering.
\label{sec:results}

\subsection{RPC: reproduction and extensions}
%! SECTION: RPC reproduces the math grid exactly; on the new domains its edge over
%! self-consistency follows the budget direction but is not significant at our sizes.
\label{sec:rpc}

%! PARA: The math grid reproduces the paper.
On the authors' reasoning paths our aggregated grid matches theirs.
Averaged over the four math benchmarks we obtain PPL $21.89/73.14$, SC $24.82/13.37$, and RPC $26.15/12.32$ (accuracy/ECE), against the paper's $21.90/73.14$, $24.82/13.37$, and $26.11/12.37$.
RPC wins on both accuracy and calibration, and PPL is accurate but badly over-confident, exactly as reported.

%! PARA: Across new domains RPC's edge is conditional.
\cref{tab:rpc-domains} extends RPC to four new domains under a small budget ($K{=}8$).
RPC's only positive movement is on KO-VER, where the model is uncertain with diverse answers and confidence tracks correctness; elsewhere it ties SC.
None of these per-item differences is statistically significant, however (paired Wilcoxon $p\ge0.28$ on every domain), so they are at most weak trends on small sets.
KCC is the informative case: the model is uncertain ($43\%$ on a four-way task, chance $25\%$) and answer-diverse, yet RPC ties SC and PPL is the best-calibrated method.
Uncertainty and answer diversity are therefore necessary but not sufficient for RPC to help; the path probabilities must also carry information about which answer is correct, which they do not for precedent grading.

\begin{table}[t]
\centering
\small
\setlength{\tabcolsep}{3pt}
\begin{tabular}{lccc}
\toprule
Domain & SC & PPL & RPC \\
\midrule
BIRD text-to-SQL & 27.5/37.7 & 26.5/72.1 & 27.0/\textbf{34.3} \\
KO-VER legal & 19.1/52.4 & 18.7/78.8 & 20.0/\textbf{46.0} \\
KCC (4-cls) & 43.1/51.1 & 42.8/\textbf{45.1} & 43.3/51.1 \\
LFUD fallacy & 88.0/12.1 & 87.0/\textbf{8.3} & 88.0/12.1 \\
\bottomrule
\end{tabular}
\caption{RPC across new domains (Qwen3-8B, $K{=}8$), accuracy/ECE ($\%$).
Bold marks the best-calibrated method (lowest ECE) per task; accuracy is left unbolded because no per-item accuracy difference is significant (paired Wilcoxon $p\ge0.28$ on every domain). PPL is best-calibrated on the closed-label tasks.
The KO-VER row covers a contiguous $150$-item block of that corpus's test split rather than a random sample of it, and its labels are citation-derived and noisy (\cref{app:kover}).}
\label{tab:rpc-domains}
\end{table}

%! PARA: On three public benchmarks RPC's edge over SC is again not significant.
To reduce reliance on our constructed datasets, we add three \emph{public} benchmarks (\cref{tab:rpc-public}): grade-school math (GSM8K~\citep{cobbe2021gsm8k}), first-order-logic entailment (FOLIO~\citep{han2024folio}), and logical-reasoning multiple choice (LogiQA~\citep{liu2020logiqa}), sampling $K{=}8$ Qwen3 paths each.
The pattern from the constructed domains holds: RPC ties SC on GSM8K (both $96.0$, near ceiling), edges it on FOLIO ($+1.1$), and trails it on LogiQA ($-2.0$), and no per-item difference is significant (Wilcoxon $p\ge0.31$).
PPL is again badly over-confident on the open-ended tasks (ECE $46$ on FOLIO, $32$ on LogiQA).

\begin{table}[t]
\centering
\small
\setlength{\tabcolsep}{3pt}
\begin{tabular}{lcccc}
\toprule
Benchmark & $n$ & SC & PPL & RPC \\
\midrule
GSM8K & 200 & 96.0/2.1 & 95.0/3.0 & 96.0/2.2 \\
FOLIO$^\dagger$ & 180 & 58.3/5.2 & 42.8/46.4 & 59.4/7.8 \\
LogiQA & 200 & 48.5/22.4 & 50.0/32.2 & 46.5/25.2 \\
\bottomrule
\end{tabular}
\caption{RPC on three \emph{public} benchmarks (Qwen3-8B, $K{=}8$), accuracy/ECE ($\%$).
The edge over SC is not significant on any (per-item Wilcoxon: GSM8K identical, FOLIO
$p{=}0.33$, LogiQA $p{=}0.31$). $^\dagger$FOLIO label parse-rate $74\%$.}
\label{tab:rpc-public}
\end{table}

%! PARA: K-scaling moves in the predicted budget direction, but the gap is small,
%! not significant, and vanishes when the sample is enlarged to n=200.
RPC trails SC on BIRD at $K{=}8$, which the original paper attributes to using a budget below its $K{=}64$--$128$ regime.
We test this directly by generating $K{=}32$ paths and re-aggregating at growing $K$ (\cref{fig:asymmetry}a, \cref{tab:rpc-kscale}).
The budget direction is as predicted: over $20$ seeds RPC goes from $26.8{\pm}0.7$ at $K{=}8$ (below SC's $27.7{\pm}0.4$) to a $+2.5$ lead at $K{=}32$, where it also nearly halves the calibration error.
The gap is small, though: at $K{=}32$ only $2$ of $80$ items separate the two methods, so a paired test does not reach significance (Wilcoxon $p{=}0.16$) and a bootstrap over items puts the $+2.5$ gap at $95\%$ CI $[0.0,+6.3]$.
On this small set the $+2.5$ is at most a weak directional signal, not an established advantage: enlarging BIRD to $n{=}200$ erases it entirely (the gap becomes $-0.25$, Wilcoxon $p{=}0.85$; see Limitations).

\begin{table}[t]
\centering
\small
\begin{tabular}{lccc}
\toprule
$K$ & SC & RPC & RPC vs SC \\
\midrule
8  & $27.7{\pm}0.4$/37.7 & $26.8{\pm}0.7$/34.3 & SC higher (n.s.) \\
16 & $28.5{\pm}0.4$/36.4 & $28.2{\pm}0.5$/31.3 & within noise \\
32 & 27.5/38.2 & 30.0/26.7 & $+2.5$ (n.s.) \\
\bottomrule
\end{tabular}
\caption{BIRD K-scaling, accuracy/ECE ($\%$). Accuracy is mean${\pm}$95\% CI over
$20$ seeds for $K{=}8,16$ (sub-sampled from $32$ cached paths); $K{=}32$ uses all
paths (single deterministic run). The budget direction is as predicted, but the
$K{=}32$ gap is not significant (per-item paired Wilcoxon $p{=}0.16$, $2/80$ items);
ECE is single-run.}
\label{tab:rpc-kscale}
\end{table}

\subsection{LCF: reproduction and critical analysis}
%! SECTION: LCF's logic-validity signal is real but weak and model-dependent, and
%! its flagship metric is unauditable.
\label{sec:lcf}

%! PARA: The headline effect is model-dependent and, on the one model it helps,
%! not statistically significant.
\cref{tab:lcf} reports our LCF reproduction across four models.
Qwen3, Llama-2, and Mistral use a single option-scoring eval so their baseline and intervention are directly comparable; Vicuna is a separate independent re-implementation, reported but not on the same scale (see caption).
The fallacy-identification $\Delta$Prob, the paper's clearest effect, rises on Qwen3-8B ($6.11\!\to\!7.83$), but a per-item paired test finds this change not significant (Wilcoxon $p{=}0.56$), and the same recipe \emph{significantly} degrades Llama-2-7b-chat ($10.50\!\to\!2.44$, $p{<}0.001$) and Mistral ($14.20\!\to\!3.23$, $p{=}0.002$); the independent Vicuna re-implementation also drops ($19.23\!\to\!18.35$ in a GB10 fp16 re-run, borderline: Wilcoxon $p{=}0.20$, $t$-test $p{=}0.04$).
On the public FOLIO logic benchmark the same recipe degrades even Qwen3, the one model it helped on the private fallacy task ($\Delta$Prob $13.58\!\to\!9.40$, $n{=}50$, not significant: $p{=}0.65$), so the lone positive effect does not carry to public logic data.
Re-training the Qwen3 projector under five seeds gives a stable $\Delta$Prob (mean $7.59$, sd $0.33$, every seed above the $6.11$ baseline), so the Qwen3 gain is reproducible across training rather than a single-run artifact, even though it stays item-concentrated and per-item non-significant.
The one positive effect is therefore both model-dependent and statistically indistinguishable from noise, while the harms are real.

\begin{table}[t]
\centering
\small
\begin{tabular}{llS[table-format=2.1]S[table-format=2.2]c}
\toprule
Model & & {Acc} & {$\Delta$Prob} & {paired $p$} \\
\midrule
\multirow{2}{*}{Qwen3-8B} & Orig & 29.9 & 6.11 & \\
                          & +LCF & 31.4 & 7.83 & 0.56 \\
\multirow{2}{*}{Llama-2-7b} & Orig & 32.4 & 10.50 & \\
                            & +LCF & 27.0 & 2.44 & $\mathbf{<.001}$ \\
\multirow{2}{*}{Mistral-7B} & Orig & 35.3 & 14.20 & \\
                            & +LCF & 27.5 & 3.23 & $\mathbf{.002}$ \\
\multirow{2}{*}{Vicuna-7b$^\dagger$} & Orig & 40.7 & 19.23 & \\
                            & +LCF & 40.2 & 18.35 & 0.20 \\
\bottomrule
\end{tabular}
\caption{LCF fallacy identification ($n{=}204$); the top three models from one
option-scoring eval, Vicuna from its independent harness ($^\dagger$).
$\Delta$Prob rises on Qwen3 but the per-item change is not significant, and drops
significantly on Llama-2 and Mistral. $p$ is a per-item paired Wilcoxon test of the
$\Delta$Prob change. $^\dagger$Vicuna is an independent re-implementation (a separate
Claude-generated validity judge), re-run here in fp16 on different hardware and a newer
transformers than the other three, so its \emph{absolute} scale is not comparable; the
degradation direction reproduces and its own per-item test is borderline (Wilcoxon
$p{=}0.20$, $t$-test $p{=}0.04$).}
\label{tab:lcf}
\end{table}

\begin{figure}[t]
\centering
\includegraphics[width=0.82\linewidth]{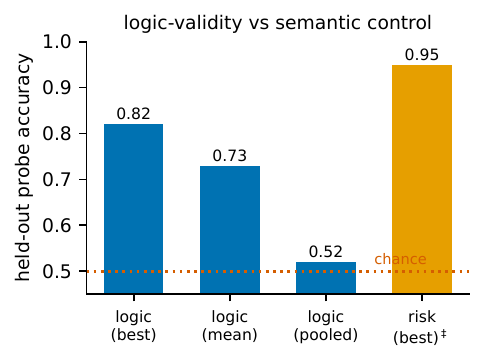}
\caption{The logic-validity direction is real but weak: a held-out probe reaches $0.82$ at the best sub-layer but $0.52$ (chance) pooled over edited layers, while a semantic attribute (suicide-risk, MoodRisk$^\ddagger$) reaches $0.95$.
$^\ddagger$control on a separate corpus.}
\label{fig:probe}
\end{figure}

%! PARA: A probe shows the logic direction is real but weak.
To explain the inconsistency we probe the representation directly (\cref{fig:probe}).
A held-out linear probe separates valid from invalid conclusions at $0.82$ accuracy at the single best sub-layer, identically for both models, which supports the paper's premise that a logic-validity direction exists.
The signal collapses to chance ($0.52$) when pooled over the layer range LCF edits.
As a control, the same probe on a semantic attribute (suicide-risk, on a mental-health corpus) reaches $0.95$ across all layers.
Representation editing works for semantic attributes; logic validity is encoded weakly and locally, which is why a recipe that mixes layers under-exploits it.

%! PARA: Separability is not controllability.
A weak-but-real direction would still be useful if shifting along it changed behavior, but in our experiments it does not.
The additive intervention, a best-layer supervised direction applied with a norm-relative shift swept over a range of strengths, gives no consistent gain on either model (the sweep is in \cref{app:cases}).
The non-additive intervention fails too: decoding the trained projector at the single best sub-layer (where the probe peaks) leaves $\Delta$Prob unchanged on both models (Qwen3 $6.11\!\to\!6.08$, paired $p{=}0.90$; Llama-2 $10.50\!\to\!10.74$, $p{=}0.40$).
Nor is the supervised direction special: a \emph{random} direction of the same norm, swept over the same $\alpha$, traces the same curve (Qwen3 $\Delta$Prob $3.96\!\to\!2.13$, Llama-2 $4.85\!\to\!1.80$ over $\alpha{=}0$--$4$, versus supervised $3.96\!\to\!2.40$ and $4.85\!\to\!1.85$), so the probe-validated logic direction buys no controllability over noise (\cref{tab:v2}).
Separability and controllability therefore come apart here: whether we add a supervised or a random direction, or decode the trained projector, a decodable logic direction is not a causal lever for logical behavior.

%! PARA: The flagship metric is unauditable.
The paper's most dramatic number, $96.56\%$ validity under a self-trained discriminator, relies on a discriminator that was not released.
Our analogue is degenerate (it marks all generations valid for one model), and the auditable GPT-4 judge shows no gain.
We find no evidence of fabrication; we find that the headline leans on an unauditable component, so it cannot be reproduced as stated.

\subsection{Toward model-agnostic logic steering}
%! SECTION: Contrastive activation steering does not make logic steering
%! model-agnostic, and we identify why.
\label{sec:agnostic}

%! PARA: We replace the trained projector with contrastive steering.
Since trained LCF is model-dependent, we test whether the activation-steering literature offers a model-agnostic alternative.
Contrastive Activation Addition (CAA)~\citep{rimsky2024steering} builds a steering vector from the mean difference of valid and invalid residual activations, with no trained module; conditional variants~\citep{lee2025programming, valentino2026mitigating} apply it only when a gate fires.
We implement a faithful K-CAST gate (kNN-based Conditional Activation Steering, following \citealp{valentino2026mitigating}: a $k$NN classifier on reference activations that decides per token whether to steer), automatic layer selection by separability, and a signed coefficient sweep.

%! PARA: On the fallacy task all variants degrade both models.
On fallacy identification every variant degrades both Qwen3 and Llama-2 (\cref{tab:steer}), and the conditional gate is inert: it fires on $98\%$ of tokens and so behaves like static steering.
The gate cannot discriminate because the reference texts (short conclusion sentences) and the task tokens come from different distributions, so a classifier that is $0.88$-separable on the references classifies nearly every task token as steer-worthy.

\begin{table}[t]
\centering
\small
\begin{tabular}{llcc}
\toprule
Task & Setting & Qwen3 & Llama-2 \\
\midrule
\multirow{2}{*}{Fallacy ($\Delta$Prob)} & original     & 3.96 & 4.87 \\
                                        & K-CAST       & 0.74 & 4.00 \\
\addlinespace
\multirow{2}{*}{Syllogism (Acc)}        & original     & 97.5 & 50.0 \\
                                        & content-abl. & \textbf{100.0} & 50.0 \\
\bottomrule
\end{tabular}
\caption{Activation steering.
On the fallacy task the conditional gate fires on $98\%$ of tokens and degrades both models.
On the matched-distribution syllogism task, ablating the content direction lifts Qwen3 to $100\%$ but cannot help Llama-2, which is at chance.
Bold marks the sole configuration that improves over its unsteered baseline.}
\label{tab:steer}
\end{table}

%! PARA: On a matched-distribution syllogism task one configuration helps.
To remove the distribution mismatch we build a controlled syllogism task crossing formal validity with believability, the design used by \citet{valentino2026mitigating}.
With reference and task drawn from the same distribution, ablating the \emph{believability} direction (not adding a validity direction) raises Qwen3 to $100\%$ and removes its content-effect gap, the first positive steering result in our study.
The win is narrow: it requires the matched distribution, the content direction rather than the validity direction, and a model that already performs the task.
Llama-2 answers a constant ``no'' to all syllogisms (chance), so no steering helps it.
A separate legal-domain generalization test is likewise inconsistent: LCF degrades Legal-LCF ($72.5\!\to\!62.5$) but improves KCC-legal ($60.0\!\to\!67.5$) on the same model. The two legal sets share only $1/40$ test premises with their training data, so this inconsistency is not a train--test leakage artifact.

\subsection{Discussion: strengths and weaknesses}
%! SECTION: We summarize the strengths and weaknesses of each method and model.
\label{sec:disc}

%! PARA: RPC's strengths and weaknesses.
\paragraph{RPC.} RPC's appeal is that it reproduces the original result exactly, trains nothing, and never significantly degraded accuracy on any domain we tested; on the original math grid it improves both accuracy and calibration, though on our new domains its edge over self-consistency does not reach significance.
Its limits follow from the same preconditions: it needs a sufficient sample budget, answer diversity, and confidence that tracks correctness, and it collapses back to plain self-consistency whenever any of these is missing, as on closed-label or already-confident tasks.

%! PARA: LCF's strengths and weaknesses.
\paragraph{LCF.} LCF rests on a premise that is genuinely real: a logic-validity direction exists, is decodable at a localized layer, and produces a real $\Delta$Prob effect on at least one model.
Its weaknesses are that this direction is weak relative to a semantic attribute, that separability does not buy controllability, that the effect transfers to no other model or domain we tried, and that the flagship validity number depends on an unreleased discriminator.

%! PARA: Model capability, not the intervention, dominates: only Qwen3 benefits
%! from LCF and from steering, and Llama-2 lacks the competence to recover.
\paragraph{Per model.} Qwen3-8B is the only model on which LCF improves $\Delta$Prob, and the only one on which contrastive steering debiases the syllogism task; it has the capability that steering can free.
Llama-2, Mistral, and Vicuna are all degraded or unaffected by LCF, and Llama-2 lacks the syllogistic competence for steering to recover.
Model capability, not the intervention, is the dominant factor.

\section{Conclusion}
%! SECTION: Test-time aggregation reproduces exactly and moves in the predicted direction
%! without reaching significance; logic-representation editing does not reproduce as a
%! reliable, model-agnostic intervention.

%! PARA: RPC's aggregation reproduces exactly and is budget-directional but not significant;
%! LCF's logic editing does not reproduce as a model-agnostic intervention.
We reproduced two reliability methods and found an asymmetry between them.
RPC's test-time aggregation reproduces the original math grid exactly and, on new domains, moves in the direction its preconditions (sufficient budget, answer diversity, informative confidence) predict, though without reaching significance at our evaluation sizes.
LCF's logic-representation editing rests on a real but weak, locally-encoded signal.
That signal is decodable yet not controllable, shows a positive effect on only one of four models that does not reach significance, significantly reduces $\Delta$Prob on two others, and leans on an unauditable metric.
Neither trained editing nor training-free contrastive steering makes it model-agnostic, except under a narrow matched-distribution condition on a model that already has the target capability.
The asymmetry rests on a narrow but real difference rather than on either method's positive gains: across every domain we tested RPC never significantly degraded accuracy (its calibration is sometimes worse than SC's, e.g.\ FOLIO and LogiQA, but we do not test ECE for significance), whereas LCF significantly reduced two of four models' $\Delta$Prob (the per-item probability margin on the valid option; the accuracy drops were not separately significance-tested), and RPC's own $+2.5$ edge on BIRD reverses to $-0.25$ once the sample grows to $n{=}200$. On this evidence, and at the $7$--$8$B model sizes and small budgets we tested, aggregation is the safer intervention, though both methods' positive effects stay non-significant.

\section*{Limitations}
%! SECTION: Beyond the methods' own weaknesses, our study is limited by small
%! evaluation sets (now with seed CIs + significance tests), re-implementation
%! fidelity, two probe confounds, and a narrow model-size range.

%! PARA: Beyond the per-method weaknesses already discussed, our study has limits
%! of evaluation size, re-implementation fidelity, and model scope.
Beyond the weaknesses of the two methods themselves (\cref{sec:disc}), our study has its own limitations.
Several extension experiments use small evaluation sets ($n{=}80$--$320$). We report $95\%$ seed confidence intervals wherever sub-sampling leaves room for them and per-item paired tests throughout (\cref{tab:lcf,tab:rpc-kscale}); these show that neither method's advantage reaches significance at these sizes, so our positive findings should be read as consistent directions rather than established effects, and larger evaluation sets would be needed to confirm them.
These sets are also statistically underpowered: a power analysis on the per-item variance shows that at $K{=}32$ on BIRD ($n{=}80$) we have $80\%$ power only for gaps $\ge 5.0$ accuracy points, well above the observed $+2.5$ (and the other domains are similar), so a non-significant result bounds the effect size rather than establishing that the effect is absent.
Expanding BIRD to $n{=}200$ probes this directly: the RPC$-$SC gap moves from $+2.5$ to $-0.25$ (Wilcoxon $p{=}0.85$), so with more data the advantage disappears rather than sharpening, consistent with the original gap being a small-sample fluctuation rather than a real effect awaiting power.
Our LCF re-implementation follows the paper but, without released code, may differ in unstated details; the substituted discriminator means our generation-validity numbers are not directly comparable to the paper's.
Two construction confounds temper the probe result: valid conclusions are model-generated (GPT-4o-mini) while invalid ones are the dataset's, so the $0.82$ separability may partly reflect surface style rather than logic (which, if anything, makes the true logic signal weaker, not stronger); the generated valid/invalid labels were not human-validated, and an independent judge (Claude) on a sample accepts only $28\%$ of the generated ``valid'' conclusions as strictly valid while agreeing the ``invalid'' ones are invalid $95\%$ of the time, so the \emph{valid} label is noisy and the logic-validity target is partly label noise (which weakens, not strengthens, our reading of the signal); and the legal corpora are Korean, so model language proficiency is a possible factor.
We evaluate Qwen3-8B, which is not one of the paper's models, and four 7--8B models in total, so our model-dependence claim is bounded to this size range.
Our conclusions about LCF are negative results under a faithful re-implementation, not proof that no implementation can reproduce the paper.

\bibliography{custom}

\appendix
\crefname{section}{Appendix}{Appendices}\Crefname{section}{Appendix}{Appendices}

\section{Dataset Construction and Examples}
%! SECTION: Per-dataset construction and one worked example each, in most
%! detail for the three corpora a reader cannot download.
\label{app:data}

\subsection{The corpora that are not public}
%! SECTION: KO-VER, its two derived legal sets, KCC, and MoodRisk cannot be
%! inspected by a reader, so we document their provenance, their label
%! derivation, the exact slice we evaluate, and what we release instead.
\label{app:kover}

%! PARA: Where KO-VER's text comes from and under what terms.
\paragraph{KO-VER: corpus and sources.} KO-VER is assembled by a crawler over Korean court decisions published on the national statute portal (\url{law.go.kr}) together with three affiliated public services covering industrial-accident, national-tax, and local-tax adjudications.
The pipeline indexes decision listings, parses the detail pages into structured records, extracts article-level statutory citations along with the sentence each occurs in, and resolves every citation to the statute \emph{version} in force at the decision date through the portal's amendment API.
The originating project reports a corpus of roughly $169{,}000$ decisions and $1.4$ million article-level citations at the time of our snapshot.
Its licensing review rests on Article~7 of the Korean Copyright Act, which places statutes and judicial decisions outside copyright protection.
Source text is released only for the operators whose open licence that review could verify.
The two tax services, whose terms it could not, ship as document identifiers plus a rebuild script.
We consume the corpus locally under those terms and redistribute none of its text.

%! PARA: What one item is, and why citation-derived labels are noisy.
\paragraph{KO-VER: items and labels.} An item is one sentence from a decision, called the context, paired with every statutory citation that sentence makes, each recorded as a $(\text{law name}, \text{article}, \text{paragraph}, \text{version key})$ tuple.
The labels are \emph{citation-derived}: they come from the hyperlinked citations in the court's own published text, not from annotators.
That is what makes them cheap at this scale, and also what makes them noisy, since a court may cite an article procedurally or in passing rather than apply it, and hyperlink extraction misses some articles outright.
The originating project measures both effects on a companion retrieval split of the same corpus: only $66.7\%$ of citation-derived pairs there score as substantively relevant under an applicability rubric, and an LLM judge recovers about half an additional applicable article per case that hyperlink extraction had missed.
Sentence-context coverage across the corpus is $74.4\%$.
Our task uses the law name and article and discards the paragraph and version fields, so the version-resolution component of the benchmark is not exercised here.

%! PARA: Exactly which items our KO-VER numbers cover: a law-based split, our
%! prefix of 150 from it, and the gold sets we ship because the corpus moves.
\paragraph{KO-VER: the split and the items we evaluate.} The benchmark splits by law name rather than by case, so that a test statute is unseen during training.
Laws are ranked by citation frequency, and of those cited at least $50$ times one in ten is assigned to test and a second, offset one in ten to validation, with every remaining law kept for training.
The assignment is made on a record's \emph{first} citation, so a test record can still cite training-split laws in its other gold pairs; the unseen-statute guarantee holds for the primary law of an item, not for its whole gold set.
We evaluate the \emph{first} $150$ records of the test split.
That is a contiguous block of the file rather than a random draw from it, so our KO-VER numbers cover the block and not the split as a whole.
Those $150$ items carry $2.52$ distinct $(\text{law}, \text{article})$ pairs on average and up to $18$, $63$ of them have exactly one pair, and $89$ distinct laws appear across their gold sets.
The corpus is maintained by a daily incremental crawl, so the test file is not frozen and its first $150$ records today no longer match the snapshot we ran in June 2026.
We therefore ship the canonical gold sets, all $K{=}8$ sampled completions, and the per-path mean log-probabilities for our exact items, which is enough to recompute every SC, PPL, and RPC number in \cref{tab:rpc-domains} without the corpus.

%! PARA: How we sampled and how a completion is scored.
\paragraph{KO-VER: generation and scoring.} We prompt Qwen3-8B in Korean for the laws and articles applied in the decision, in a \mbox{``$\langle$law name$\rangle$ 제N조''} list after an \texttt{Answer:} marker, with paragraph and subparagraph explicitly excluded and one two-line format example.
Generation is bfloat16 with thinking mode disabled, temperature $0.8$, top-$p$ $0.95$, at most $256$ new tokens, contexts truncated to $4{,}000$ characters, and $K{=}8$ samples per item.
A completion is parsed by locating every \mbox{``$\langle$law name$\rangle$ 제N조[의M]''} span.
Law names are compared with all whitespace removed, so a spaced and an unspaced writing of the same decree match while the suffix separating an enforcement decree from its parent act survives.
Article numbers are re-emitted canonically from their digits.
A sample counts as correct only when its parsed set equals the gold set exactly, and an empty parse never matches, including against another empty parse; every one of the $1{,}200$ sampled completions parsed to at least one pair.
Set-exact scoring against a mean of $2.52$ pairs is a demanding criterion.
It is the main reason all three methods sit near $20\%$ on this domain, and since they consume identical paths, the comparison between them is unaffected.
Label noise and set-exact scoring do cost statistical power, though.
The non-significant RPC--SC difference on KO-VER (\cref{tab:rpc-domains}) therefore bounds the effect on this block rather than showing that aggregation cannot help on legal citation extraction, and a cleaner label set could resolve a gap this one cannot.

%! PARA: How the two derived legal LCF sets were generated from those corpora.
\paragraph{Legal-LCF and KCC-legal.} Both derived sets follow the LFUD data contract so the LCF pipeline consumes them unchanged.
For Legal-LCF we take KO-VER contexts, keep one context per case so that a scenario is a case, skip contexts under $60$ characters, truncate the rest to $1{,}200$, and stop at $200$ cases.
One GPT-4o-mini call per premise, under a Korean system prompt, returns both conclusions.
The invalid one targets a named legal fallacy on the same facts: over-generalising a holding, affirming the consequent of a statutory condition, conflating a necessary with a sufficient condition, or ignoring a stated exception.
Scenarios are split $70{:}10{:}20$ under seed $42$, giving $140/20/40$.
The four-option identification item holds the valid conclusion, this case's invalid one, one invalid conclusion drawn from another case in the same split, and a Korean ``cannot be determined on these facts'' option; the gold index is the valid conclusion.
KCC-legal is built identically from KCC precedent holdings.
Both members of a legal pair are generated, where LFUD supplies its own fallacious member, so the surface-style confound we raise for LFUD is smaller here while neither member is corpus-grounded; no lawyer reviewed either set.

%! PARA: The remaining two unavailable corpora and what we use them for.
\paragraph{KCC and MoodRisk.} KCC is a Korean civil precedent corpus of $2{,}939$ query--candidate pairs with graded $0$--$3$ relevance labels, accepted for publication and released with its own paper~\citep{cho2026kcc}.
We draw a class-balanced subset of $80$ pairs per grade, and release that subset's items together with our sampled paths.
MoodRisk is a mental-health corpus of $8{,}736$ annotated social-media posts, also under preparation, which we use only as a probe control on mean-pooled layer representations to bound how strongly a semantic attribute is linearly encoded (\cref{fig:probe}).
No MoodRisk text or representation is redistributed here; we release the probe script and its outputs.

\subsection{Worked examples}
%! SECTION: One input-output example per constructed dataset.

%! PARA: One input->output example per constructed dataset; Korean shown verbatim.
To make each constructed dataset concrete, one input$\rightarrow$output example per dataset is given below.
Korean-language data (KO-VER, KCC, Legal-LCF) is shown in the original, untranslated; long fields are truncated with ``[\dots]''.

%! PARA: LFUD conclusion-generation example: dataset fallacy as invalid, GPT-built valid.
\paragraph{LFUD conclusion generation (LCF).} The invalid conclusion is LFUD's
fallacy; the valid one is generated with GPT-4o-mini.\\
\textbf{Input (premise):} All electronic products need electricity.\\
\textbf{Output (valid):} Some electronic products require electricity to function.\\
\textbf{Output (invalid):} All electronic products need electricity.
Electronics are part of my family.
Therefore, everything in my house needs electricity.

%! PARA: LFUD fallacy-identification example: pick the option committing the fallacy.
\paragraph{LFUD fallacy identification (RPC).} Four candidate arguments; the model
picks the one committing the stated fallacy.\\
\textbf{Input:} \emph{Faulty generalization} draws a conclusion about all instances from a few.
(0)~Most people think all flowers do not stay open forever, so it must be true.
(1)~Gardeners, experts on flowers, claim all flowers do not stay open forever, so it is so.
(2)~All flowers do not stay open forever; roses are plants; so all plants do not stay open forever.
(3)~Nobody can show a flower that stays open
forever, so none do.\\
\textbf{Output:} (2).

%! PARA: BIRD example: question -> gold SQL, scored by execution match.
\paragraph{BIRD text-to-SQL (RPC).} Scored by execution match against the gold query.\\
\textbf{Input (question):} What is the highest eligible free rate for K-12 students
in the schools in Alameda County?\\
\textbf{Output (gold SQL):} {\small\texttt{SELECT \dots FROM frpm WHERE County Name = 'Alameda' ORDER BY \dots\ DESC LIMIT 1}}

%! PARA: KO-VER example: decision context -> set of (law, article) pairs.
\paragraph{KO-VER citation extraction (RPC; Korean).} \textbf{Input (context):} 건설업법 시행령 제3조 는 건설업 면허기준은 기술능력 자본금 시설건설 영업세 납부액 및 건설업 경력이 영업의 종류에 따라 건설업 면허 기준표의 기준에 이르러야 한다고
하였고 [\dots]\\
\textbf{Output (law/article set):} \{(건설업법시행령, 제3조)\}

%! PARA: KCC example: query + candidate precedent -> graded relevance (0-3).
\paragraph{KCC precedent grading (RPC; Korean).} \textbf{Input (질의):} 사건명: 손해배상청구사건. 판시사항: 도급인의 손해배상책임이 발생하기 위한 요건. 판결요지: 도급인이 특정한 행위를 지휘하거나 [\dots] 민법 제756조 소정의 사용자의
배상책임 규정에 의한 책임이 있다.\\
\textbf{Input (후보):} 사건명: 손해배상(기). 판시사항: [1] 도급인이 수급인에 대하여 특정한 행위를 지휘하거나 특정한 사업을 도급시키는 노무도급의 경우, 사용관계가 인정되는지 여부(적극)
[\dots]\\
\textbf{Output (grade):} 2 (상당히 관련 / substantially relevant).

%! PARA: Legal-LCF example: statute premise -> valid vs invalid Korean conclusion.
\paragraph{Legal-LCF conclusion pairs (LCF; Korean).} \textbf{Input (전제):} 건설업법 시행령 제3조 는 [\dots] 여기에서 자본금이라 하는 개념은
건설업에 실질적으로 사용될 수 있는 재산을 가리킨다.\\
\textbf{Output (valid):} 건설업 면허의 자본금 기준을 충족하기 위해서는 건설업에 실질적으로
사용될 수 있는 재산만이 포함되어야 한다.\\
\textbf{Output (invalid):} 건설업 면허를 취득하기 위해서는 모든 형태의 재산이 자본금으로 인정될 수 있다.

%! PARA: Syllogism example: two premises -> Yes/No; full 2x2 design in the table.
\paragraph{Syllogism (steering).} Two premises and a conclusion; output Yes if the conclusion formally follows, else No. The full validity$\times$believability 2$\times$2 design is in \cref{tab:syll}.

\begin{table}[h]
\centering
\small
\begin{tabular}{llp{4.7cm}}
\toprule
Cell & Ans & Premises $\to$ conclusion \\
\midrule
VB & Yes & All flowers are plants; all roses are flowers $\to$ all roses are plants \\
VU & Yes & All flowers are roses; all plants are flowers $\to$ all plants are roses \\
IB & No  & All roses are plants; all flowers are plants $\to$ all roses are flowers \\
IU & No  & All roses are plants; all flowers are plants $\to$ all flowers are roses \\
\bottomrule
\end{tabular}
\caption{Syllogism 2$\times$2: V/I = formally valid/invalid, B/U = believable/ unbelievable.
The conflict cells (VU, IB) isolate the content effect.}
\label{tab:syll}
\end{table}

\section{Worked Cases}
%! SECTION: Worked cases behind the RPC and LCF claims.
\label{app:cases}

%! PARA: A KO-VER case where RPC beats SC because confidence tracks correctness.
\paragraph{When RPC beats self-consistency.} \cref{tab:rpc-case} shows a real KO-VER item.
The gold answer is a single statute (Civil Procedure Act \S139).
Most sampled paths append a spurious Civil-Act article, so the majority vote (SC) returns a wrong superset; but the lowest-perplexity paths ($\text{lp}=-0.12$) return the clean correct answer, so perplexity-weighting (PPL/RPC) selects it.
This is the regime where confidence tracks correctness.

\begin{table}[h]
\centering
\small
\begin{tabular}{clp{3.6cm}}
\toprule
path & mean lp & predicted statutes \\
\midrule
0 & $-0.17$ & CivAct \S224 \& CivProc \S139 \\
1 & $-0.16$ & CivAct \S242 \& CivProc \S139 \\
2 & $\mathbf{-0.12}$ & \textbf{CivProc \S139} (correct) \\
4 & $\mathbf{-0.12}$ & \textbf{CivProc \S139} (correct) \\
7 & $-0.12$ & CivAct \S225 \& CivProc \S139 \\
\bottomrule
\end{tabular}
\caption{KO-VER item (5 of $K{=}8$ paths).
SC majority returns a wrong superset (CivAct \S242 \& CivProc \S139); PPL/RPC select the lowest-perplexity clean answer (CivProc \S139 = gold).}
\label{tab:rpc-case}
\end{table}

%! PARA: The LCF v2 strength sweep that shows separability is not controllability.
\paragraph{LCF v2 strength sweep.} \cref{tab:v2} is the sweep behind \cref{sec:lcf}: shifting along the best-layer supervised logic direction (probe accuracy $0.818$) by a norm-relative coefficient $\alpha$ gives no consistent $\Delta$Prob gain on either model, and degrades both at larger $\alpha$.

\begin{table}[h]
\centering
\small
\begin{tabular}{cS[table-format=2.1]S[table-format=1.2]S[table-format=2.1]S[table-format=1.2]}
\toprule
$\alpha$ & \multicolumn{2}{c}{Qwen3-8B} & \multicolumn{2}{c}{Llama-2-7b} \\
 & {Acc} & {$\Delta$Prob} & {Acc} & {$\Delta$Prob} \\
\midrule
0.0 & 31.9 & 3.96 & 39.2 & 4.85 \\
0.5 & 30.4 & 3.93 & 40.2 & 4.71 \\
1.0 & 29.9 & 3.91 & 39.2 & 4.35 \\
2.0 & 30.9 & 4.06 & 32.8 & 3.37 \\
4.0 & 28.4 & 2.40 & 30.4 & 1.85 \\
\bottomrule
\end{tabular}
\caption{LCF v2 best-layer steering sweep (supervised logic direction). No consistent
gain at any $\alpha$; separability ($0.818$) does not yield controllability, and a
matched-norm \emph{random} direction traces the same curve (\cref{sec:lcf}), so the
direction is not special. $\Delta$Prob here is from the additive-sweep harness, whose
option scoring differs from \cref{tab:lcf}, so the $\alpha{=}0$ baseline is not
numerically identical to the projector eval there.}
\label{tab:v2}
\end{table}

\end{document}